\documentclass[conference]{IEEEtran}
\IEEEoverridecommandlockouts
\usepackage{cite}
\usepackage{amsmath,amssymb,amsfonts}
\usepackage{algorithmic}
\usepackage{graphicx}
\usepackage{textcomp}
\usepackage{xcolor}
\usepackage{booktabs}

\def\BibTeX{{\rm B\kern-.05em{\sc i\kern-.025em b}\kern-.08em
    T\kern-.1667em\lower.7ex\hbox{E}\kern-.125emX}}
\begin{document}

\title{ACDZero: MCTS Agent for Mastering Automated Cyber Defense}

\author{
\IEEEauthorblockN{
Yu Li\textsuperscript{\textdagger,1}, 
Sizhe Tang\textsuperscript{\textdagger,1}, 
Rongqian Chen\textsuperscript{1},
Fei Xu Yu\textsuperscript{1}, 
Guangyu Jiang\textsuperscript{1}, 
Mahdi Imani\textsuperscript{2},
Nathaniel D. Bastian\textsuperscript{3},
Tian Lan\textsuperscript{1}}
\IEEEauthorblockA{
\textsuperscript{1}Dept of ECE, George Washington University, Washington, D.C., USA\\
\textsuperscript{2}Dept of ECE, Northeastern University, Boston, MA, USA\\
\textsuperscript{3}Dept of EECS, United States Military Academy, West Point, NY, USA\\
}\\

\thanks{\textsuperscript{\textdagger}Equal contribution.}
}
\newcommand{\n}{\textsc{ACDZero }}

\maketitle

\begin{abstract}
Automated cyber defense (ACD) seeks to protect computer networks with minimal or no human intervention, reacting to intrusions by taking corrective actions such as isolating hosts, resetting services, deploying decoys, or updating access controls. However, existing approaches for ACD, such as deep reinforcement learning (RL), often face difficult exploration in complex networks with large decision/state spaces and thus require an expensive amount of samples. Inspired by the need to learn sample-efficient defense policies, we frame ACD in CAGE Challenge 4 (CAGE-4 / CC4) as a context-based partially observable Markov decision problem and propose a planning-centric defense policy based on Monte Carlo Tree Search (MCTS). It explicitly models the exploration-exploitation tradeoff in ACD and uses statistical sampling to guide exploration and decision making. We make novel use of graph neural networks (GNNs) to embed observations from the network as attributed graphs, to enable permutation-invariant reasoning over hosts and their relationships.
To make our solution practical in complex search spaces, we guide MCTS with learned graph embeddings and priors over graph-edit actions, combining model-free generalization and policy distillation with look-ahead planning. We evaluate the resulting agent on CC4 scenarios involving diverse network structures and adversary behaviors, and show that our search-guided, graph-embedding-based planning improves defense reward and robustness relative to state-of-the-art RL baselines.
\end{abstract}


\section{Introduction}

\begin{figure*}[htbp]
    \centering \includegraphics[width=0.95\linewidth]{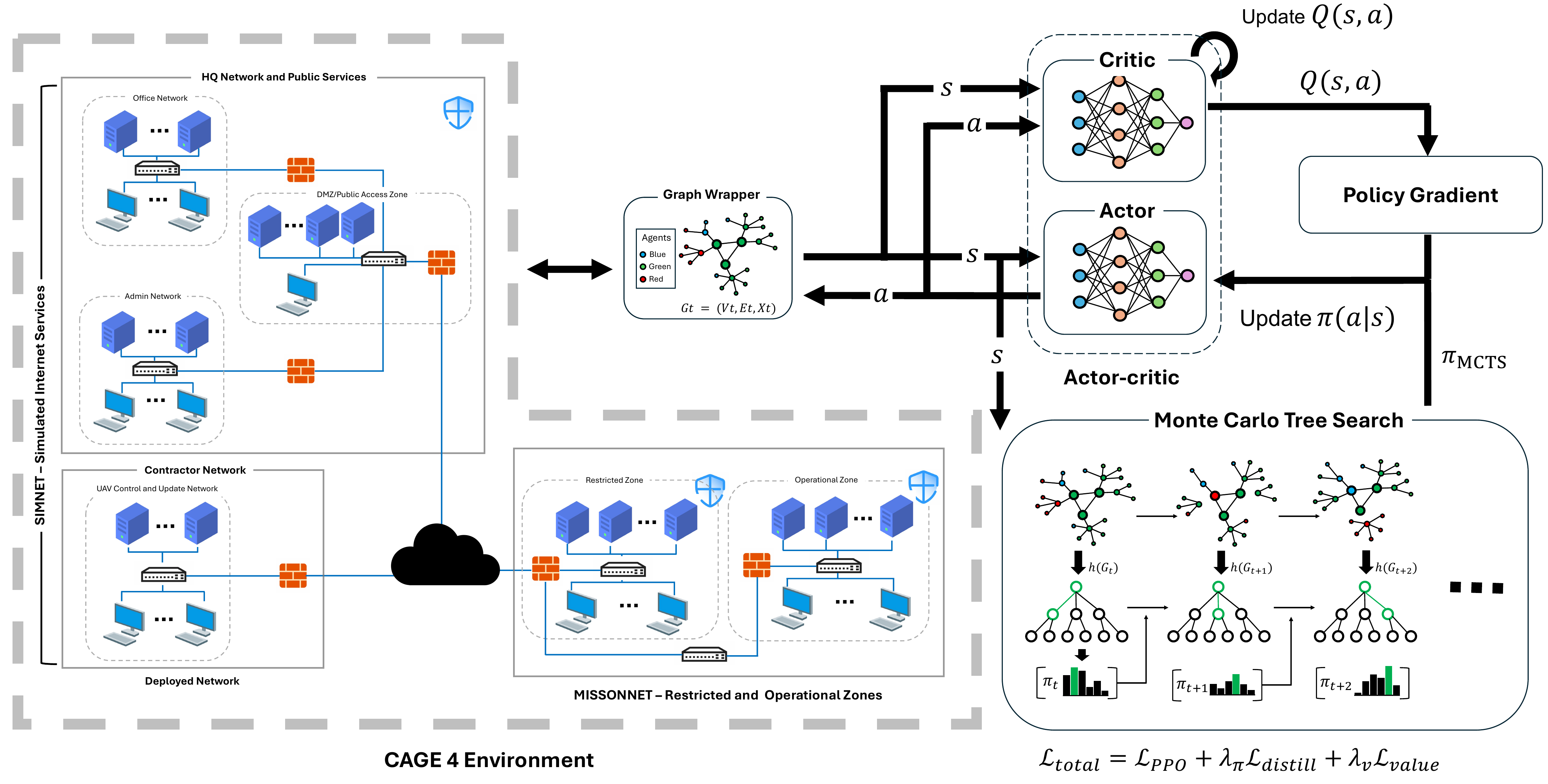}
    \caption{Overview of the ACDZero framework applied to the CAGE Challenge 4 environment. CAGE-4 simulates a high-fidelity enterprise network where autonomous defenders must protect critical assets against adaptive adversaries in a partially observable, multi-agent setting. To address the brittleness of standard reactive policies that degrade under topological changes, our approach treats defense not only as policy learning but as online decision-time planning. The system first transforms local observations into attributed graphs ($G_t$) to enable permutation-invariant reasoning over hosts and their relationships. Uniquely, ACDZero utilizes a Monte Carlo Tree Search (MCTS) module (bottom right) to perform look-ahead planning within a learned latent space, simulating alternative futures to discover optimal strategies. These high-quality search policies serve as distillation targets for the GNN-based Actor (top right), allowing the agent to internalize strategic foresight while maintaining the inference speed required for real-time autonomous cyber defense.
    }
    \label{fig:overview}
\end{figure*}

Automated cyber defense (ACD) systems are designed to monitor network environments and execute corrective actions—such as host isolation, service restoration, decoy deployment, and credential rotation—with minimal human input~\cite{vyas2023automated, tang2025human}. While reinforcement learning (RL) and deep reinforcement learning (DRL) have demonstrated potential in training defense policies within simulated environments~\cite{nguyen2021deep, tang2023edge}, the approaches often face the challenge of balancing exploration and exploitation~\cite{DBLP:conf/icml/DamPDW25,10.1007/978-981-96-6599-0_2,li2025inspo}
in complex networks with large decision/state spaces, and thus require an expensive amount of samples to learn a reasonable defense policy. As modern cyber-adversaries become more sophisticated~\cite{chen2025neurosymbolic, chen2025perception, hong2025poster}  and leverage multi-step strategies. Existing solutions relying on RL and DRL struggle to capture multi-step look-ahead defense planning while also meeting the required sample efficiency for agile cyber defense.

To bridge this gap, we propose \textbf{\textsc{ACDZero}}, a framework for learning automated cyber defense policies through Monte Carlo Tree Search (MCTS) and graph-based latent-space planning.
More precisely, we leverage MCTS, which has demonstrated strong performance in multi-step reasoning and complex planning problems~\cite{swiechowski2023monte} such as mastering chess and board games~\cite{muzero,DBLP:conf/ecai/CzechBK24,silver2017masteringchessshogiselfplay}-to explicitly model the exploration-exploitation tradeoff in ACD and use statistical sampling to guide decision making. By building a dynamic model of the ACD problem (in a latent graph-embedding space as introduced later), each search consists of a series of simulated ACD games of defense action self-play to traverse a tree from root state to leaf state of the ACD game. Each simulation proceeds by selecting in each node/state a defense action with respect to a dynamically-updated upper confidence tree (UCT) bound to balance exploration-exploitation. The final game result of each rollout is then used to weight the nodes in the ACD game tree and to update the UCT bounds for learning a search policy with optimal rewards. 

Unique challenges arise from applying MCTS to a complex ACD environment like the CAGE Challenge 4 (CC4)~\cite{kiely2025cage}, involving multiple subnets, dynamic communications, random initialization, and a large set of servers/hosts and defense actions. We note that structural heterogeneity and unknown environment dynamics due to partial observability lead to significant challenges in representing the dynamic observations and network states to learn a dynamic model in MCTS. This dynamism and uncertainty render standard fixed-length vector representations brittle and prone to failure when deployed across varying network topologies~\cite{vargas2021review, ccavucsouglu2024novel,li2025dual,li2025sfmdiffusion,zeng2024local, li2024crowdsensing}. 
To this end, \n makes novel use of a Graph Neural Network (GNN) as an invariant structural engine to encode network entities as typed nodes and edges~\cite{guan2024graph, yan2023swingnn}. This representation serves as the foundation for a learned latent dynamics model, which enables MCTS to perform "virtual" look-ahead simulations without access to the ground-truth simulator~\cite{muzero}.

Crucially, our architecture integrates this high-fidelity search into a decentralized Actor-Critic framework. During training, the MCTS serves as a powerful policy improvement operator, generating strategic targets that are distilled into a GNN-based actor via policy gradient updates. This distillation process allows the agent to internalize complex multi-step reasoning within its neural weights. At deployment, the distilled policy can be executed directly as a fast, reactive actor, retaining the strategic foresight of search-guided training without the computational overhead of real-time planning. \n is implemented in the CC4 environment and evaluated against state-of-the-art RL baselines.

Our contributions are summarized as follows:
\begin{itemize}
    \item \textbf{ACDZero Framework:} We introduce a graph-embedding-based MCTS framework that unifies GNN-based state representation with a learned latent dynamics model and an Actor-Critic distillation pipeline for topology-robust planning.
    
    \item \textbf{Formalization of Graph-Based POMDP:} We formalize ACD in CC4 as a partially observable Markov decision process over dynamic graphs, necessitating representations that are strictly invariant to node permutation and network scale.
    
    \item \textbf{State-of-the-Art Performance:} We demonstrate that \n achieves a 29.2\% improvement in defense success over the current state-of-the-art graph-based baselines, while exhibiting superior convergence stability across diverse and unseen network configurations.
    
    \item \textbf{Empirical Analysis of Search-Guided Training:} Through extensive ablation studies, we quantify the contributions of latent-space planning and policy distillation, confirming that MCTS-guided supervision is the primary driver for achieving strategic robustness in complex cyber environments.
\end{itemize}

\section{Backgrounds}

\textbf{Automated Cyber Defense.}
ACD has transitioned from traditional rule-based heuristics ~\cite{vyas2023automated} toward RL frameworks capable of discovering optimal defensive policies through autonomous environment interaction~\cite{nguyen2021deep, hammar2020finding, kunz2022multiagent, jiang2022intelligent}. However, the efficacy of RL in network security is often constrained by state representation. Early approaches predominantly utilized fixed-vector encodings, where network features are mapped to static indices~\cite{ridley2018machine, kiely2309autonomous, miehling2015optimal, xu2023cnn}. Such representations impose an implicit dependency on node ordering; because they lack permutation invariance, even a minor re-indexing of hosts results in disparate feature vectors for functionally identical network states, severely inhibiting generalization.

These architectural limitations are particularly pronounced in high-fidelity benchmarks such as the CC4 ~\cite{kiely2025cage}. Unlike static environments, CC4 features a stochastically initialized topology where the number of hosts (5–15 per subnet) and active services (1–5 per host) vary across episodes. This structural fluidity prevents agents from memorizing specific configurations and demands strategies that are robust to topological shifts. To mitigate this, recent research has pivoted toward graph-based representations~\cite{king2025automated, collyer2022acd,chang2025calibrating}. By encoding the network as an attributed graph—where entities are nodes and relations are edges—defenders can leverage GNNs to achieve the permutation invariance necessary for reasoning about structural patterns across heterogeneous network configurations~\cite{guan2024graph}.

\textbf{Monte Carlo Tree Search.} 
MCTS is applied widely to solve planning problems through sequential decision-making ~\cite{mcts_survey, tangmalinzero}. A typical MCTS involves four phases, i.e., \textit{Selection} to choose actions from candidates via UCB-style strategies \cite{ucb1}, \textit{Expansion} to sample new candidate actions for existing nodes, \textit{Simulation} to obtain the corresponding payoffs, and \textit{Backup} to update the cumulated returns along the search path. We denote the state by \(s\) and action by \(a\). For each node \((s,a)\) in the tree, there are statistics including the estimated value \(\Phi(s,a)\), visiting count \(N(s,a)\).

MuZero~\cite{muzero} is a classic MCTS framework learning an internal dynamics model, allowing it to perform tree-based planning without access to the environment's ground-truth rules or a simulator. This framework is composed of three learnable components parameterized by $\theta$: (i) a representation function $s_0 = h_\theta(o_1, \dots, o_t)$ that transforms observations into a latent state, (ii) a dynamics function $(s_k, r_k) = g_\theta(s_{k-1}, a_k)$ that predicts the next latent state and immediate reward, and (iii) a prediction function $\left(\mathbf{p}_k, v_k\right) = f_\theta(s_k)$ which outputs the policy prior and state value. 

During the MCTS process, MuZero navigates the search tree entirely within this learned latent space. Starting from the root node, it applies a variant of the predictor Upper Confidence Bound (pUCT) \cite{muzero} to select actions:
\(a = \arg\max_{a\in\mathcal{A}} Q(s,a) + P(s,a) \frac{\sqrt{\sum_b N(s,b)}}{N(s,a)+1} c_1\)
where $P(s,a)$ is the prior probability from the policy head and \(c_1\) is a constant. 
Statistics along the search path are updated during the \textit{Backup} phase using a cumulative discounted payoff $G_{t,k} = \sum_{\tau=0}^{l-1-k} \gamma^\tau \hat{r}_{k+1+\tau} + \gamma^{l-k} v^l$. The mean action value $Q(s,a)$ is then updated as the average of these bootstrapped returns: $Q(s,a) = \frac{N(s,a)Q(s,a) + G_{t,k}}{N(s,a) + 1}$.

\section{method}

\subsection{Problem Formulation}
\label{subsec:problem_formulation}

We formalize the automated cyber defense task within the CC4 networks environment as a Decentralized Partially Observable Markov Decision Process.
Therefore the problem can be defined by the tuple $\mathcal{M} = \langle \mathcal{N}, \mathcal{S}, \mathcal{A}, \mathcal{O}, \mathcal{T}, \mathcal{R}, \gamma \rangle$. 
Here, $\mathcal{N}$ denotes the set of defender agents, $\mathcal{S}$ the global state space encompassing network topology and host compromise status, $\mathcal{A}$ the joint action space, and $\mathcal{O}$ the observation space. 
The transition function $\mathcal{T}: \mathcal{S} \times \mathcal{A} \rightarrow \Delta(\mathcal{S})$ and reward function $\mathcal{R}: \mathcal{S} \times \mathcal{A} \rightarrow \mathbb{R}$ capture the dynamics arising from interactions among red (attacker), green (user), and blue (defender) agents.

Due to partial observability, each agent $i \in \mathcal{N}$ receives only a local observation $o^{(i)}_t \in \mathcal{O}$ corresponding to its assigned subnet.
Therefore, a key challenge is that the network topology is stochastically initialized at each episode, causing the dimensionality of $\mathcal{S}$ and $\mathcal{O}$ to vary. 
To address this, we frame policy learning over dynamic graphs rather than fixed-dimensional vectors, requiring representations invariant to input size and node permutation.

\subsection{State Representation and Environment Interface}
\label{subsec:state_representation}

To bridge CC4 simulation data and the graph-based policy architecture, we implement a specialized \textit{Environment Interface} that (i) constructs semantically rich attributed graphs from local observations, and (ii) maps policy decisions to executable simulation commands.

At each timestep $t$, the interface transforms the agent's local observation into an attributed graph $\mathcal{G}_t = (\mathcal{V}_t, \mathcal{E}_t, \mathbf{X}_t)$, where $\mathcal{V}_t$ denotes the node set, $\mathcal{E}_t$ the edge set, and $\mathbf{X}_t \in \mathbb{R}^{|\mathcal{V}_t| \times d}$ the $d$-dimensional node feature matrix. 
We represent environment entities as typed nodes: \textit{Hosts} (servers and workstations), \textit{Subnets} (network segments), \textit{Ports} (services and connections), and \textit{Files} (analyzed assets). Each node $v \in \mathcal{V}_t$ encodes type-specific attributes---hosts include OS metadata (version, distribution, architecture) and role indicators; ports encode process information (port number, service type) and status flags (ephemeral, default, decoy); files capture analysis metadata (density, signature status). Categorical attributes are one-hot encoded.

A critical component is inter-agent communication integration. CC4 restricts communication to 8-bit messages, which the interface parses and encodes as features of corresponding \textit{Subnet} nodes, enabling implicit coordination without centralized training. Environment-wide variables are encoded into a global context vector $\mathbf{g}_t \in \mathbb{R}^{d_g}$ for temporal and phase awareness.
For action mapping, the interface abstracts simulator actions into graph operations: defensive operations (\texttt{Analyze}, \texttt{Restore}, \texttt{DeployDecoy}) target \textit{Host} nodes, while network operations (\texttt{AllowTraffic}, \texttt{BlockTraffic}) modify edges between subnet nodes. This decoupling enables seamless adaptation to variable network topologies.

\subsection{\n}
\label{subsec:acdzero_framework}

\begin{figure}[t]
    \centering \includegraphics[width=1.0\linewidth]{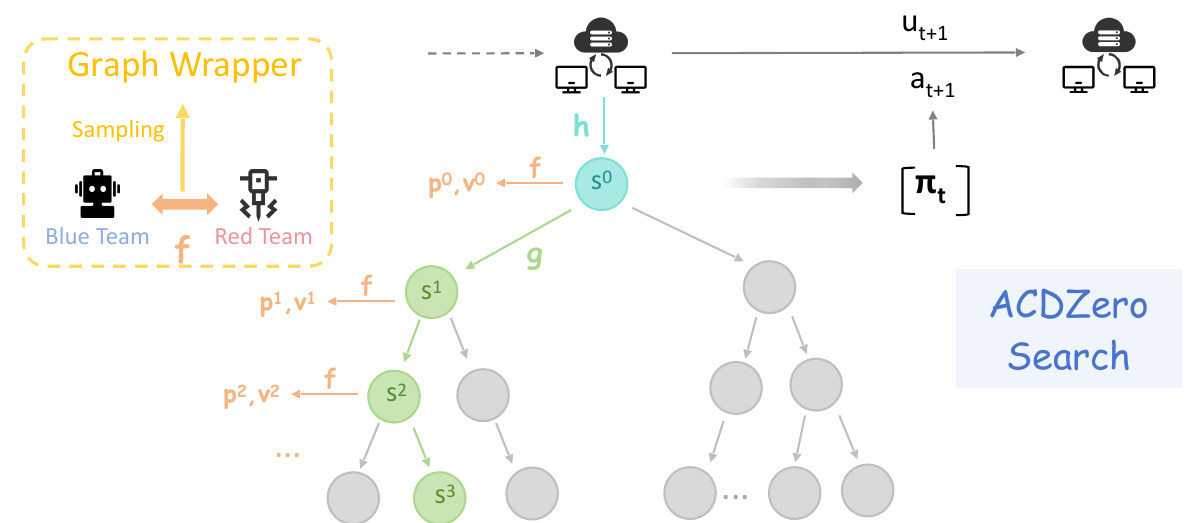}
    \caption{\n~framework. Graph observations are encoded into latent states. MCTS performs tree search with learned dynamics $g$ to generate improved policy and value estimates $(p, v)$, which are distilled via prediction head $f$ into the actor $\pi_\theta$ for action selection.}
    \label{fig:placeholder}
\end{figure}

Our proposed algorithm, \n, adopts a MuZero-like framework \cite{muzero} for decentralized cyber defense. The core objective is to employ MCTS as a policy improvement operator that enhances a GNN-based PPO \cite{ppo} agent. MCTS performs lookahead search in learned latent space, generating an improved policy $\pi_{\text{mcts}}$ and value estimates that are distilled into decentralized actor-critic networks. 

In our \n, network state evolution is modeled as transitions in a latent search tree. For each agent $i$, search initializes at a root node representing the latent belief $s_{t,0} = h_\theta(o_{\le t})$, where $h_\theta$ is a GNN-based representation network processing graph-structured observation history $o_{\le t}$.
To capture dependencies in stochastically initialized CC4 topologies, $h_\theta$ employs two-stage hierarchical aggregation: 
\textit{Intra-Entity Aggregation} pools port and file attributes into host embeddings $\mathbf{h}_{host}$, 
then \textit{Inter-Subnet Aggregation} propagates these to subnet nodes, creating a representation invariant to node permutations and network size.
Tree edges represent defensive actions $a \in \mathcal{A}^{(i)}$, and child nodes are latent states predicted by dynamics function $s_{t,k+1} = g_\theta(s_{t,k}, a_{t+k})$. To handle variable action dimensionality, $g_\theta$ projects each action into fixed-dimensional embeddings processed by a GRU, capturing temporal dependencies across heterogeneous topologies. The algorithm integrates a reward function $r_{t,k} = R_\theta(s_{t,k}, a_{t+k})$ and prediction head $(\mathbf{p}_{t,k}, v_{t,k}) = f_\theta(s_{t,k})$ to forecast rewards, policy priors, and state values.

\paragraph{MCTS Procedure}
To perform multi-step reasoning in CC4's latent space, each agent executes a fixed number of look-ahead simulations before taking real actions, mimicking rehearsal of attack-defense trajectories. The procedure follows three iterative phases:

\begin{itemize}
    \item \textbf{Selection:} Starting from root node $s_{t,0}$, the agent traverses the tree by selecting actions balancing exploitation and exploration. We employ the pUCT rule \cite{muzero} with Dirichlet noise. At each node $s$, the agent selects action $a^*$ according to:
    \begin{equation}
        a^* = \arg\max_{a \in \mathcal{A}^{(i)}} \left[ Q(s, a) + P(s, a) \cdot \frac{\sqrt{\sum_{b} N(s, b)}}{1 + N(s, a)} c_1 \right]
    \end{equation}
    where $Q(s, a)$ tracks the action's historical performance, and $P(s, a)$ from prediction head $f_\theta$ represents the agent's prior intuition. Selection continues until reaching a leaf node.

    \item \textbf{Expansion and Evaluation:} At leaf node $s_{t,l}$, \n~performs virtual expansion using learned dynamics. Unlike traditional MCTS requiring a simulator, \n~uses dynamics function $g_\theta$ to generate the next latent state $s_{t,l+1} = g_\theta(s_{t,l}, a_{t+l})$ and predicts reward $\hat{r}_{t,l}$. Simultaneously, prediction head $f_\theta$ evaluates the node to obtain value $v_{t,l+1}$ and policy prior $\mathbf{p}_{t,l+1}$, enabling anticipation of defensive operations without environment interaction latency.

    \item \textbf{Backup:} Evaluation results propagate backwards to update ancestor node statistics. For each state-action pair $(s, a)$, we increment visit count $N(s, a)$ and update mean value $Q(s, a)$ using $n$-step bootstrapped return $G_{t,k}$:
    \begin{equation}
    G_{t,k} = \sum_{j=0}^{l-k-1} \gamma^j \hat{r}_{t,k+j} + \gamma^{l-k} v_{t,l},
    \end{equation}
    \begin{equation}
    Q(s, a) \leftarrow \frac{N(s, a)Q(s, a) + G_{t,k}}{N(s, a) + 1}
    \end{equation}
    where $v_{t,l}$ is the terminal value and $\hat{r}$ are rewards predicted by $R_\theta$. This recursive update ensures root statistics converge toward an optimal defensive strategy, providing robust targets for policy distillation.
\end{itemize}

\paragraph{Optimization Objectives.} 
The training \n~is formulated as multi-task learning, integrating MCTS's deliberate reasoning with PPO's reactive efficiency. Each agent minimizes a joint loss, ensuring stable updates, effective distillation, and accurate latent dynamics:
\begin{equation}
    \mathcal{L}_{total} = \mathcal{L}_{PPO} + \lambda_{\pi} \mathcal{L}_{distill} + \lambda_{v} \mathcal{L}_{value}
\end{equation}
where $\lambda_{\pi}$ and $\lambda_{v}$ are scaling coefficients that balance the contribution of each objective.

\begin{itemize}
    \item \textbf{Decentralized Policy Optimization ($\mathcal{L}_{PPO}$):} To maintain baseline stability in the non-stationary multi-agent environment, we employ the standard clipped surrogate objective:
    \begin{equation}
        \mathcal{L}_{PPO} = \mathbb{E}_{t} \left[ \min(r_t(\theta) \hat{A}_t, \text{clip}(r_t(\theta), 1-\epsilon, 1+\epsilon) \hat{A}_t) \right]
    \end{equation}
    where $r_t(\theta) = \frac{\pi_\theta(a_t|o_t)}{\pi_{\theta_{old}}(a_t|o_t)}$ is the probability ratio and $\hat{A}_t$ is the advantage estimated by the GNN-critic. This loss ensures that the agent's policy does not deviate excessively from its previous iterations, facilitating safe exploration.

    \item \textbf{MCTS-Guided Policy Distillation ($\mathcal{L}_{distill}$):} A pivotal feature of \n is the use of MCTS as a "policy improvement" operator. The search process yields a visit count distribution at the root node, which constitutes an improved search policy $\pi_{\text{mcts}}(a|s) \propto N(s, a)^{1/\tau}$. We use the Kullback-Leibler (KL) divergence to force the GNN actor $\pi_\theta$ to internalize the multi-step look-ahead logic:
    \begin{equation}
        \mathcal{L}_{distill} = \mathbb{E}_{t} \left[ D_{KL}(\pi_{\text{mcts}}(\cdot | s_t) || \pi_{\theta}(\cdot | o_t)) \right]
    \end{equation}
    By minimizing this loss, the lightweight actor learns to approximate the high-fidelity search policy, allowing it to exhibit "strategic foresight" even during real-time inference when the search tree is omitted.

    \item \textbf{Latent Dynamics and Value Prediction ($\mathcal{L}_{value}$):} To ensure the latent space provides a reliable foundation for planning, the dynamics function $g_\theta$, the reward head $\hat{r}$, and the value head $v$ are optimized to minimize the prediction error over an unrolled trajectory of length $K$. This process facilitates \textit{indirect optimization}: the dynamics model is not supervised by raw state observations but is instead shaped by its utility in predicting rewards and long-term values. This ensures that the latent transitions capture the most semantically relevant features for cyber defense, such as host compromise status and subnet connectivity.
\end{itemize}

Through joint optimization, \n~bridges slow deliberation and fast execution. During training, MCTS supervises the GNN to learn complex defensive patterns; during deployment, the agent maintains neural network inference speed, achieving robustness against high-velocity cyber-attacks.

\begin{table*}[htbp]
\centering
\caption{Performance and interpretable cybersecurity metrics on CAGE Challenge 4.}
\label{tab:cc4_comprehensive}
\resizebox{\textwidth}{!}{
\begin{tabular}{lccccccc}
\toprule
\textbf{Method} & \textbf{Reward} & \textbf{Clean Hosts} & \textbf{Non-Escalated} & \textbf{Recovery Prec.} & \textbf{Mean TTR} & \textbf{Impact Count} & \textbf{Recovery Error} \\
& (mean±std) & (ratio) & (ratio) & (TP/TP+FP) & (timesteps) & (per episode) & (\%) \\
\midrule
DQN (Tabular)    & $-606.20 \pm 43.22$ & 0.19 & 0.82 & 0.12 & 142.3 & 9.84 & 88 \\
PPO (Tabular)    & $-597.28 \pm 41.98$ & 0.21 & 0.84 & 0.14 & 138.6 & 9.51 & 86 \\
\addlinespace[0.1cm]
GCN              & $-193.68 \pm 21.07$ & 0.74 & 0.96 & 0.61 & 58.7 & 2.45 & 39 \\
\textbf{ACDZero} & $\mathbf{-150.03} \pm \mathbf{19.85}$ & \textbf{0.82} & \textbf{0.98} & \textbf{0.71} & \textbf{46.2} & \textbf{1.28} & \textbf{32} \\
\bottomrule
\end{tabular}
}
\end{table*}

\section{Results}

We evaluate ACDZero on the CAGE Challenge 4 environment, comparing it against tabular RL baselines (DQN, PPO) and the graph-based GCN method. 
Our experiments demonstrate that combining MCTS-guided planning with graph neural networks yields substantial improvements in both final performance and sample efficiency.

\subsection{Experimental Setting}
All methods are evaluated on CAGE Challenge 4, a multi-agent cyber defense scenario where five blue agents defend against adaptive red adversaries across four network zones. 
The network topology is stochastically initialized at each episode, with 5-15 hosts per subnet and 1-5 services per host. Following the official protocol, we evaluate against FiniteStateRedAgent over 100 episodes of 500 timesteps each, reporting mean reward and standard deviation.

We compare against: (1) DQN and PPO using fixed-vector representations via the EnterpriseMAE wrapper, and (2) GCN, a graph-based method using graph convolutional networks, as they have announced, ranked 5th on the official CAGE-4 leaderboard.

ACDZero uses a GNN backbone with 256-dimensional hidden layers and 128-dimensional embeddings.
MCTS performs 16 simulations per action with dynamic $c_1$ scheduling ($c_{\text{base}} = 19652$, $c_{\text{init}} = 1.25$).
During training, we apply Dirichlet noise ($\alpha = 0.3$, $\epsilon = 0.25$) and use temperature $\tau = 1.0$; during evaluation, $\tau = 0.1$. 
The joint loss uses $\lambda_\pi = 0.5$ and $\lambda_v = 0.5$. We train with 5 parallel workers using PPO clipping $\epsilon = 0.2$ and discount factor $\gamma = 0.99$.

\subsection{Main Result}

Table~\ref{tab:cc4_comprehensive} presents the final performance of all methods. 
ACDZero achieves a mean reward of $-150.03 \pm 19.85$, representing a 29.2\% improvement over the GCN baseline ($-193.68 \pm 21.07$). 
The improvement demonstrates the effectiveness of integrating MCTS-guided planning with graph-based policy learning, which stems from MCTS systematically exploring multi-step defensive strategies, policy distillation providing high-quality training targets, and learned dynamics enabling anticipation of attacker behavior.
DQN and PPO obtain mean rewards of $-606.20$ and $-597.28$, respectively, highlighting the fundamental limitation of fixed-vector representations: they cannot generalize across varying network topologies because they implicitly memorize specific node orderings.

Beyond final performance, ACDZero exhibits 5.8\% lower variance ($\pm 19.85$ vs $\pm 21.07$) than GCN, indicating more consistent defense across diverse network configurations. The MCTS stable planning framework is adaptable to different topologies, and the learned strategies capture generalizable defense principles rather than topology-specific heuristics.

\begin{figure}[htbp]
\centering
\includegraphics[width=1.0\linewidth]{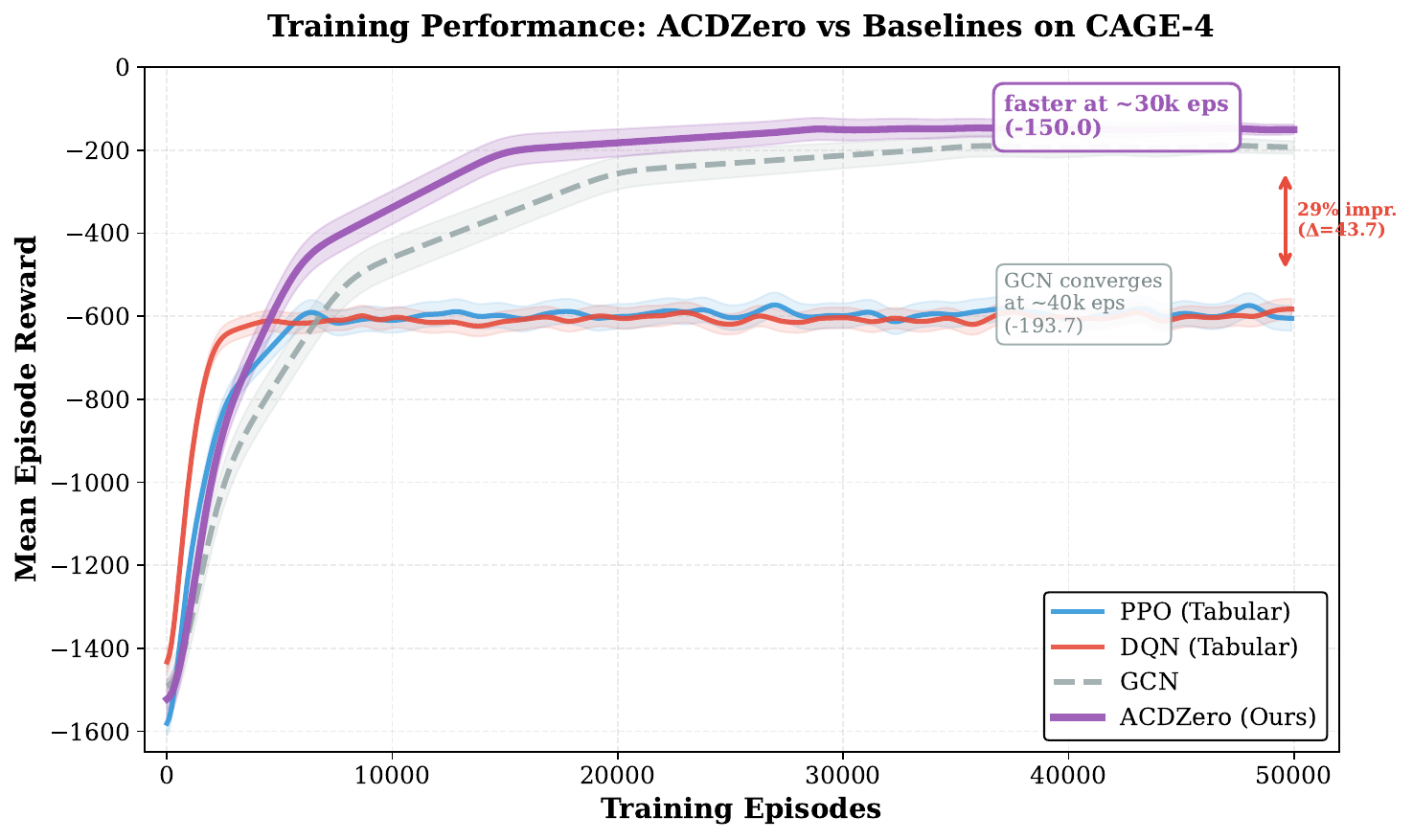}
\caption{Training performance on CAGE Challenge 4. ACDZero converges 
faster ($\sim$30k episodes) and to better performance ($-150$) than the GCN 
baseline ($-193.68$ at $\sim$40k episodes). Tabular methods plateau early at suboptimal performance. Shaded regions indicate 
standard deviation.}
\label{fig:training_curves}
\end{figure}

Figure~\ref{fig:training_curves} shows learning curves over 50,000 episodes. 
Tabular methods plateau within 10,000 episodes at $-600$, while GCN continues improving until 40,000 episodes ($-193.68$), demonstrating that permutation-invariant processing extracts topology-independent strategies.
ACDZero converges at 30,000 episodes (25\% faster than GCN) with superior performance. 
This acceleration stems from MCTS providing multi-step search supervision rather than noisy single-step rewards.
Training requires 2.5$\times$ more computation per step than GCN, but fewer episodes yield comparable total time.
At inference, ACDZero uses only $\pi_\theta$ without MCTS, achieving reactive-speed decision-making.

\subsection{Ablation Study}

To isolate the contributions of individual components, we systematically remove key elements of ACDZero. Table II summarizes the results.

Removing MCTS (using only the GNN-based PPO agent) reduces performance to $-193.68$, equivalent to the GCN baseline. This 29.2\% performance gap directly quantifies the benefit of search-guided planning. Disabling policy distillation while retaining MCTS yields $-175.23 \pm 23.41$, demonstrating that knowledge transfer from search to policy network provides substantial gains beyond using MCTS for action selection alone. Removing Dirichlet noise results in $-162.45 \pm 20.72$, showing that stochastic root exploration is important for discovering diverse defensive strategies. Using fixed $c_1 = 1.25$ instead of dynamic scheduling yields $-158.91 \pm 21.33$, indicating that adaptive exploration control provides modest but consistent improvements.

\begin{table}[t]
\centering
\caption{Ablation study on ACDZero components.}
\label{tab:ablation}
\begin{tabular}{lcc}
\toprule
\textbf{Configuration} & \textbf{Mean Reward} & \textbf{Std.} \\
\midrule
ACDZero (Full)           & $-150.03$ & $\pm 19.85$ \\
\addlinespace[0.05cm]
~~w/o MCTS               & $-193.68$ & $\pm 21.07$ \\
~~w/o Policy Distill     & $-175.23$ & $\pm 23.41$ \\
~~w/o Dirichlet Noise    & $-162.45$ & $\pm 20.72$ \\
~~w/o Dynamic $c_1$      & $-158.91$ & $\pm 21.33$ \\
\bottomrule
\end{tabular}
\end{table}

\section{Conclusion}
We presented ACDZero, a graph-guided planning framework combining graph neural networks with Monte Carlo Tree Search for automated cyber defense. 
ACDZero addresses topology generalization and multi-step reasoning challenges in dynamic network environments. By combining graph neural networks' representational flexibility with tree search's deliberative reasoning, ACDZero achieves robust performance across diverse configurations while maintaining computational efficiency for real-time deployment. Evaluation on CAGE Challenge 4 demonstrates 29.2\% performance improvement over the state-of-the-art GCN baseline, with 25\% faster convergence and 5.8\% lower variance. 
Looking ahead, \n~method enables two promising extensions: (1) integrating with pre-trained policies to leverage domain knowledge as MCTS priors, accelerating exploration and convergence, and (2) learning the graph-based dynamics model from offline trajectories collected by existing systems, such as rule-based defenders or learned baselines, thereby reducing online interaction costs while preserving adaptive planning capabilities.

\bibliographystyle{IEEEtran}  
\bibliography{ref}

\end{document}